\documentclass{article}

\usepackage{microtype}
\usepackage{graphicx}
\usepackage{subfigure}
\usepackage{booktabs} 
\usepackage[utf8]{inputenc} 
\usepackage[T1]{fontenc}    
\usepackage{hyperref}       
\usepackage{url}            
\usepackage{booktabs}       
\usepackage{amsfonts}       
\usepackage{nicefrac}       
\usepackage{microtype}      
\usepackage{amsmath,graphicx}
\usepackage{amssymb}
\usepackage{amsthm}
\usepackage{enumitem}
\usepackage{overpic}
\usepackage{color}
\usepackage{tikz}
\usepackage{wrapfig}
\usepackage{adjustbox}
\usepackage{enumerate}
\usepackage{verbatim}
\usetikzlibrary{shapes.multipart,arrows,positioning,calc}

\usepackage{hyperref}

\usepackage[accepted]{icml2018}

\newcommand {\R} {{\mathbb{R}}}
\newcommand {\N} {{\mathbb{N}}}
\newcommand {\Z} {{\mathbb{Z}}}
\renewcommand {\AA} {{\mathcal{A}}}
\newcommand {\RR} {{\mathcal{R}}}
\renewcommand {\SS} {{\mathcal{S}}}
\newcommand {\ED} {{\mathrm{ED}}}
\newcommand {\piexp}{{\pi_{\mathrm{expl}}}}
\newcommand {\piexe}{{\pi_{\mathrm{exec}}}}

\icmltitlerunning{Planning with Arithmetic and Geometric Attributes}

\begin{document}

\twocolumn[
\icmltitle{Planning with Arithmetic and Geometric Attributes}



\icmlsetsymbol{equal}{*}

\begin{icmlauthorlist}
\icmlauthor{David Folqué}{to}
\icmlauthor{Sainbayar Sukhbaatar}{to,goo}
\icmlauthor{Arthur Szlam}{goo}
\icmlauthor{Joan Bruna}{to}
\end{icmlauthorlist}

\icmlaffiliation{to}{Courant Institute of Mathematical Sciences, New York University}
\icmlaffiliation{goo}{Facebook AI Research}

\icmlcorrespondingauthor{Joan Bruna}{bruna@cims.nyu.edu}

\icmlkeywords{Machine Learning, ICML}

\vskip 0.3in
]



\printAffiliationsAndNotice{\icmlEqualContribution} 

\begin{abstract}
A desirable property of an intelligent agent is its ability to understand its environment to quickly generalize to novel tasks and compose simpler tasks into more complex ones.  If the environment has geometric or arithmetic structure, the agent should exploit these for faster generalization.   Building on recent work that augments the environment with user-specified attributes, we show that further equipping these attributes with the appropriate geometric and arithmetic structure brings substantial gains in sample complexity.
\end{abstract}

\section{Introduction}
We consider an agent in a Markovian environment that is partially specified by a set of human-defined attributes. The attributes provide a useful language for communicating tasks to the agent.   

In \cite{zhang2018composable} it was shown that using (state, attribute) pairs as supervision, it is possible for an agent to explore the environment and accomplish tasks at test time that are described by those attributes.   However, in that work, the attributes were binary functions, and the worst case complexity of exploring the attributes scaled exponentially in the number of attributes. 

Many environments of interest have geometric and/or arithmetic structure.  In this work we
show that equipping these attributes with the appropriate geometric and arithmetic structure brings substantial gains in sample complexity. We demonstrate our model on 2d grid-world environments.



\section{Problem Setup}

We start with a Markovian Enviroment (ME) $(\mathcal{S}, \mathcal{A}, P)$, given 
by a state space $\mathcal{S}$, an action space $\mathcal{A}$ and 
a transition kernel $\mathcal{S} \times \mathcal{A} \to \mathcal{S}$ 
specified by the probability $P(s' | a, s)$ to transition from state $s\in \mathcal{S}$ to $s' \in \mathcal{S}$ by taking action $a \in \mathcal{A}$. 
Model-based approaches attempt to estimate the transition kernel in order to perform planning. 
In this context, it is crucial to exploit regularity priors: in many practical scenarios this ME is highly structured, in the sense that the transition kernel varies smoothly with respect to specific transformations in the state/action spaces. For example, 
applying a force $a=\vec{F}$ to an object at location $\vec{p}$ will likely produce the same effect $\vec{p'}$ 
than applying the same action to the same object at a different location: $P(\vec{p'} | \vec{p}, \vec{F}) \approx P(\vec{p'}+\vec{p_0} | \vec{p}+\vec{p_0}, \vec{F})$. 

For that purpose, the ME is augmented with a \emph{structured attribute space} $\mathcal{R}$ and a deterministic mapping $f : \mathcal{S} \to \mathcal{R}$, encoding the attributes 
$\rho = f(s)$ associated to each state. This mapping may be either given by the user, 
or may be regressed from a dataset of labeled pairs $\{(s_i, \rho_i)\}_i$, resulting in an estimate $\hat{f}$.
Unless otherwise specified, in the following we shall write $f$ to denote the ground-truth state-attribute 
mapping. In order to leverage the regularity of the environment, we equip the attribute space $\RR$ with predefined algebraic and geometric structure. In this work, we consider attribute spaces built as outer products of elementary groups and monoids\footnote{ A monoid is a semigroup with identity element; a semigroup is a set with an associative binary operation.}, such as real numbers $\mathbb{R}$, integers $\mathbb{Z}$, counts $\mathbb{N}$ and modular arithmetic $\mathbb{Z}/(q \mathbb{Z})$. 
Our model-based approach thus amounts to estimating the  transition kernel induced in the attribute space. 
At test time, the agent will be given attribute goals $\rho_g$, and its objective is to take an appropriate sequence of actions in the original environment to reach $\rho_g$.

\section{Related work}

This work builds upon the unstructured attribute planning model from \cite{zhang2018composable}.  In that work, a Markovian state space was augmented with a set of binary attributes.  The attributes were used as a means of organizing exploration and communicating target states.  In that work, the agent was  built from three components:
(i) a neural-net based {\bf attribute detector} $\hat{f}$, which maps states $s$ to a set of attributes $\rho$, i.e. $\rho=\hat{f}(s)$.
(ii) a neural net-based {\bf policy $\pi(s,\rho_g)$} which takes a pair of inputs: the current state $s$ and attributes of an (intermediate) goal state $\rho_g$, and outputs a distribution over actions.
(iii) a {\bf transition table} $c_\pi(\rho_i, \rho_j)$ that records the empirical probability that $\pi(s_{\rho_i}, \rho_j)$ can succeed at transiting successfully from $\rho_i$ to $\rho_j$ in a small number of steps.

In this work, in addition to allowing binary attributes, we consider attributes with more algebraic structure.  In addition, we augment the transition table $c_{\pi}$ with a parametric edge detector that takes into account the structure of the attributes.
We refer the reader to the references in \cite{zhang2018composable} for a more complete review of the literature this work is built upon; but will briefly highlight a a few especially relevant works.
Because we add further structure to the attributes, this work moves the unstructured attribute planner closer to \citep{Hernandez-GardiolK03,Otterlo_Relational_RL, Diuk_object,AbelHBBOMT15}, which discuss MDPs that can be written in terms of objects and relations between those objects.  However, this current work still focuses on the interface between the symbolic description of the underlying Markovian space and the actual space; and the symbolic description in terms of attributes with algebraic structure is an approximation.






\section{The Structured Attribute Model}

In this section we describe our Structured Attribute Model, 
depicted in Figure \ref{fig:diagram}.
It contains several modules that interact with each other. 
We describe each of these modules in detail and their interactions. 

\tikzset{
block/.style = {draw, fill=white, rectangle, minimum height=3em, minimum width=3em},
tmp/.style  = {coordinate}, 
sum/.style= {draw, fill=white, circle, node distance=1cm},
input/.style = {coordinate},
output/.style= {coordinate},
pinstyle/.style = {pin edge={to-,thin,black}
}
}

    

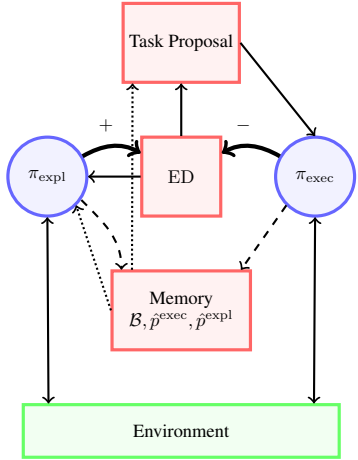
\begin{figure}
    \centering
\begin{tikzpicture}[thick,scale=0.7, every node/.style={transform shape},
roundnode/.style={circle, draw=blue!60, fill=blue!5, very thick, minimum size=15mm},
squarednode/.style={rectangle, draw=red!60, fill=red!5, very thick, minimum size=15mm},
rectnode/.style={rectangle, draw=green!60, fill=green!5, very thick, minimum width=6cm,minimum height=1cm},
]
\node[squarednode]      (edgedetector)                              {$\ED$};
\node[squarednode]      (taskprop)          [above=of edgedetector]    {Task Proposal};
\node[squarednode]      (memory)          [below=of edgedetector]    {\begin{tabular}{c} Memory \\ $\mathcal{B},\hat{p}^\mathrm{exec},\hat{p}^\mathrm{expl}$ \end{tabular} };
\node[rectnode]      (environment)          [below=of memory]    {Environment};
\node[roundnode]        (exploration)       [left=of edgedetector] {$\piexp$};
\node[roundnode]        (execution)       [right=of edgedetector] {$\piexe$};

\draw[thick, ->] (edgedetector.north) -- (taskprop.south)  ;
\draw[thick,densely dotted, ->] ([xshift=4mm]memory.north west) -- ([xshift=2mm]taskprop.south west);
\draw[thick, ->] (taskprop.east) -- (execution.north);
\draw[ultra thick, ->] (execution) node [text width=3cm,right,above=2em ] {$-$} to  [out=150,in=30](edgedetector)  ; 
\draw[thick,dashed, ->] (execution.south west) -- ([xshift=-2mm]memory.north east);
\draw[thick, <->] (execution.south) -- ([xshift=-5mm]environment.north east);
\draw[ultra thick, ->] (exploration) node [text width=3cm,midway,above=2em ] {$+$} to [out=30,in=150] (edgedetector)  ;
\draw[thick, ->] (edgedetector.west) to [out=180,in=0]  (exploration.east);
\draw[thick,dashed, ->] ([xshift=1mm,yshift=1mm]exploration.south east) to [out=315,in=90]  ([xshift=2mm]memory.north west);
\draw[thick,densely dotted, ->] (memory.west) --  (exploration.south east);
\draw[thick, <->] (exploration.south) -- ([xshift=5mm]environment.north west);

\end{tikzpicture}    
    \caption{Block Diagram of our Structurd Attribute Model. Dashed lines correspond to Memory writes, dotted lines to memory reads, and thick lines provide labels to the Edge detector. Our model thus combines a non-parametric component given by the memory, and a parametric Edge Detector that extends the agent experience through the structured attribute space.}
    \label{fig:diagram}
\end{figure}


We consider attribute spaces $\RR$ built as direct products 
of group building blocks. In this work, we consider natural arithmetic attributes $\N$, 
modular arithmetic $\Z/(q \Z)$, and real-valued attributes $\R$. We note however that our methodology can be easily extended to more exotic algebraic and geometric structures, 
such as modular real-valued attributes $S^1$, rigid motions $SO(2)$ or dihedral groups.

For the purposes of this work we need a notion of \emph{difference} vector in the attribute space. We denote by
$\delta \RR = \{ \rho_1 + ( \rho_0)^{-1} ; \,\exists s, s' \in \SS\,s.t. ~\rho_1=f(s'), \rho_0=f(s)\,,\,\sum_{a} P(s'|s,a) >0 \}~ $
the set of admissible transitions in attribute space, 
and where the group operation is taken coordinate-wise.




\subsection{Edge Detector}

The core component of our model is a module that, given a pair $(\rho, \delta\rho) \in \RR \times \delta\RR$, 
evaluates when a given transition $\rho \to \rho':=\rho + (\delta\rho)$ is feasible in the environment using at most $M_{\text{max}}$ actions, that is, whether 
$\exists  m \leq M_{\text{max}}, s_0=s, s_1, \dots, s_m=s' \in \SS \,;\, f(s) = \rho, f(s') = \rho' \,, \sum_{a_0, \dots, a_{m-1}} \prod_{l=1}^m P( s_l | s_{l-1}, a_l) > 0~.$


This network receives as input $\rho$ using the corresponding group attribute parametrisations, and 
$\delta \rho \in \RR - \RR$, 
and outputs $\ED(\rho, \delta\rho)$, the estimated 
probability that this transition is feasible. 
This detector is trained in a supervised fashion by receiving both positive and negative samples. 
The positive samples are fed by the exploration policy $\piexp$, 
whereas the negative samples are produced by the execution policy $\piexe$. 

\subsection{Memory}

Our model contains two policies, detailed next: the exploration policy and the execution policy. 
In each case, we record the empirical counts on which attributes they have visited. 
Since the size of the attribute space grows exponentially with respect to the number of 
attributes, we consider only the marginal counts. For each attribute dimension $k \leq K$, 
we consider empirical marginal counts $\hat{p}^{\mathrm{exec}}_k$, $\hat{p}^{\mathrm{expl}}_k$ over $G_k$. 
In case some attributes are continuous, we quantize them using a predefined number of bins in order to produce the empirical counts. 
Finally, in order to keep track of the admissible transitions in the full attribute space (without marginalization), we consider a memory buffer $\mathcal{B}$
that contains every observed transition $\delta \rho \in \delta\RR$, this time without marginalization. 


\subsection{Exploration Policy}

The estimation of the transition kernel starts with an exploration policy $\piexp$ that scans 
for transitions in attribute space $(\rho_i, \rho_j)$. This policy is parametrised by 
a neural network that takes as input the pair $(s_i, \rho_i=f(s_i))$ and outputs a 
distribution over $\AA$. Its rewards are determined from the Edge Detector 
and from the empirical marginal counts $\hat{p}^{\mathrm{expl}}$ as follows. 
If $\ED(\rho_i, \delta\rho)$ is the current estimated probability that the transition $\rho_i \to \rho_i + (\delta\rho)$ 
is feasible, then $\piexp$ gets rewarded when he finds an actual transition with low estimated probability using at most $M_{\text{max}}$ steps. Additionally, we reward the exploration for uncovering unseen attribute values:
  $  R_{\mathrm{exp}} = -\alpha_1 \log \ED(\rho_i, \delta\rho) \cdot \mathbf{1}[ \rho_{i+m} = \rho_i + (\delta \rho)] + \alpha_2 g(\rho_{i+m},\rho_i) ~,$
with $m \leq M_{\text{max}} $.
Here, the function $g(\rho_{i+m}, \rho_{i})$ is inversely proportional to the times the exploration has previously seen this transition, measured according to each coordinate:
$h(\rho_1, \rho_2) := \min_{k\leq K}  \mathbf{1}(\rho_1[k] \neq \rho_2[k]) \cdot p^{\mathrm{expl}}_k(\rho_1[k])^{-0.5}~.$
The weighting parameters $\alpha_1, \alpha_2$ are adjustable to each environment and are reported in the experimental section.
The episodes start where the last episode ends, and the game restarts when the agent is unable to perform more transitions. A game is always played only by one policy, either the exploration or the execution one, and the training phase consists in alternating games of both. 







\subsection{Execution Policy}

The execution policy $\piexe$ takes as input the current state-attribute pair $(s, \rho=f(s))$ as well as a target transition $\delta\rho$ (which may or may not be in $\delta \RR$) provided by the Transition Proposal module, and outputs a distribution over $\AA$. This policy is trained with reinforcement learning, via a positive reward $R_0$ whenever it reaches a 
state $s'$ such that $f(s') = \rho + \delta\rho$. If after a certain number of steps $M_{\text{max}}'$ it has failed 
to reach the desired attribute transition, it sends the sample $(\rho, \delta \rho)$ back to the 
Edge Detector with a negative label.

\subsection{ Transition Proposals}

Finally, we describe the module that proposes which transitions the execution policy should be trained on. 
We want to accomplish two objectives: (i) enforce that the coverage of the execution policy $\piexe$ matches that of the exploration $\piexp$, and (ii) provide the Edge Detector with negative samples, i.e. transitions that are not admissible in the environment. 
We propose to sample the target transitions $\delta \rho$  
 from the current buffer $\mathcal{B}$ of recorded transitions as follows.
First, we filter out the transitions $b \in \mathcal{B}$ that are considered unlikely to exist according to the current edge detector using a threshold $\ED(\rho, b)>p_0$ (we pick $p_0 = 0.1$ in all experiments); call $\mathcal{B}^+$ the remaining transitions. 
Then we consider a mixture that samples uniformly at random within $\mathcal{B}^+$ with probability $s_0$, 
and according to the following distribution with probability $1-s_0$: 
$\forall\, \delta\rho \in \mathcal{B}^+~,~p( \delta \rho  | \rho) = \frac{\gamma(\delta \rho)}{\sum_{b \in \mathcal{B}^+} \gamma(\delta \rho)}~,$ where $~ \gamma(\delta \rho) = \frac{\min_k \hat{p}^{\mathrm{expl}}_k((\rho+\delta\rho)[k])}{1+\min_k \hat{p}_k^{\mathrm{exec}}((\rho+\delta\rho)[k])}~.$
In words, we look at the differences in marginal counts between $\hat{p}^{\mathrm{expl}}_k$ and $\hat{p}^{\mathrm{exec}}_k$ 
across the recorded transitions in the buffer, and sample more often  those where exploration outpaces execution.

\subsection{Inference}
\label{inferenceplanning}

Our inference strategy at test time is analogous to the unstructured attribute work \cite{zhang2018composable}, 
except that our estimated probabilities to realize each transition are given by the Edge Detector. 
Specifically, we look for the path $(\rho_0,\rho_1,\dots,\rho_K)$ from the start $\rho_0=f(s)$ to our goal $g=\rho_K$ that minimizes the distance defined by the Edge Detector as: $d(a,b)=-\log(\ED(a,b-a))$.
To do this we use Dijkstra's algorithm on the graph that starts at the point where the agent is and extends to other points in the attribute space by applying the transitions in the buffer, giving each edge $(p,q)$ the cost $-\log(\ED(p,q-p))$.

\section{Experiments}


We report preliminary experiments on Grid-World games using 
the Mazebase environments \cite{sukhbaatar2015mazebase}. It consists of 2-D maps that vary dynamically at each episode, of size between $7$ and $10$ in each dimension, with a single agent that interacts with the environment.
In all scenarios, we train our structured model without access to the test-time tasks during a prespecified 
number of episodes. After training, given a target task, we perform planning using Djikstra as explained in Section \ref{inferenceplanning}.
For simplicity, we assume the state-to-attribute mapping $\rho=f(s)$ is known in our reported experiments. 
We consider two baselines: 
    (i) A reinforcement learning agent parametrised with the same neural network architecture as our execution policy, taking the state and goal attribute as inputs, 
trained using a curriculum that starts from nearby tasks and extends them to the evaluation goals, and (ii) the Unstructred Attribute Planner from \cite{zhang2018composable}. Here, we treat attributes as a set. When the environment contains continuous attributes, we round them to the nearest integer and use the resulting discrete space.  





\begin{table*}
\footnotesize
\centering
\begin{tabular}{ |c|ccc| } 
 \hline
  & Modular Switches & Exchangeable Attr & Constrained Attr \\ \hline
 RL+curriculum & 13.6\% & 14\% & 0.2\% \\ 
 Unstructured Attribute Model & 9.6\% & 88\% & 20.8\% \\
 Structured Attribute Model & \bf{89.3\%} & \bf{93\%} & \bf{81.6\%} \\ 
 \hline
\end{tabular}
\caption{Percentage of proposed tasks that have been successfully accomplished using a fixed budget of allowed steps on three different Grid-World Environments. We consider a budget of $150$ steps for all models. In the RL setting, we train the model on the same conditions as faced during testing. In the attribute setting, all the training is agnostic to the tasks proposed at test-time.}
\label{mazebaseresults}
\end{table*}

\subsection{Modular Switches}

The first environment consists in 2-d mazes containing a variety of different objects. Depending on the state of a switch, the agent is allowed to pick objects of a specific kind, as illustrated in  Figure \ref{fig:games}. 
 For each object type, we consider two attributes: how many objects are still available in the map, and how many objects the agent already collected. The attribute space is thus modeled with 
$\RR = \underbrace{\mathbb{N} \times \dots \mathbb{N}}_{2 q \text{ times}} \times (\mathbb{Z}/q \mathbb{Z})~,$
where $q$ corresponds to the number of different objects. We consider $q=3$ in our experiment.
This environment is highly structured, and the only admissible transitions are of the form 
$\delta \rho = (e_j; 0)$ with $e_j[j]=1, e_j[j+q]=-1$, $j \leq q$, and
$\delta \rho = (0; 1)$. 
The evaluation tasks consist in requesting a specific number of items of each caterogy $\rho_{\mathrm{target}}=(n_1, \dots, n_q)$. The RL version is trained with a curriculum that grows the distance (induced by the transitions in the attribute space) between the start and the end of the task, from 1 to 15.
In this environment, Table \ref{mazebaseresults} shows that neither the curriculum RL agent nor the Unstructured Attribute Planner model are able to successfully complete the target goals. The number of transitions is large relative to the number of steps. 
RL trained in 50M steps, and our model as well as the Unstructured Attribute is trained in 25M steps for exploration and 25M steps to train the policy. Figure \ref{fig:labelcounts} displays the positive (resp. negative) transitions discovered by $\piexp$ (resp. $\piexe$) as training progresses. 
Our model is able to quickly generalize the transition kernel of the environment to unseen regions, by leveraging its rich arithmetic structure. 


\subsection{Exchangeable Attributes}

This environment contains objects of several types, and for each type we consider two attributes: how many objects are still in the map, and how many objects the agent has already collected. 
At any time, the agent has the possibility to trade objects using pre-specified exchange rates, as shown in Figure \ref{fig:games}. 
The attributes of this environment can be modeled as 
$\RR = \underbrace{\mathbb{N} \times \dots \mathbb{N}}_{2 q \text{ times}}~,$
but this time the admissible transitions create interactions between attributes. The exchange rates determine  transitions of the form $(\rho[i], \rho[j]) \mapsto (\rho[i]-e_{i,j}, \rho[j] + f_{i,j})$, for $i, j =1\dots q$. Inspired by real markets, we set $e_{i,j} > f_{i,j}$ for all pairs. 
The evaluation tasks consist in obtaining a predefined number of items of each type. We consider as before the case  $q=3$. The RL with curriculum is  trained in 40M steps. Our model trained in 20M steps for exploration and 20M steps to train the policy. Similarly as before, the RL curriculum is implemented by growing the distance (induced by the transitions in the attribute space) between the start and the end of the task, from 1 to 30. 
In this case, Table \ref{mazebaseresults} shows that, while our structured attribute model still outperforms the two baselines, the difference is less dramatic than in other environments. We attribute this to the fact that the transition kernel is more homogeneous and faster mixing than before, and therefore although the exploration in the unstructured attributes misses many transitions, the planning phase manages to cover the attribute space more efficiently than in the modular switch environment.


\subsection{Constrained Attributes}

Finally, we consider an environment with a continuous attribute component. 
Here, an agent is deployed in an terrain collecting minerals with a single-use hammer. Each time the hammer is used, the agent receives a random amount of mineral coming from a distribution over $\mathbb{R}_{+}$. The agent can go to the `hardware store' and obtain a new hammer in order to keep mining. He can also go to the dump yard and throw away a fixed amount of mineral. 
In that case, the attribute space is modeled as 
$\RR = \underbrace{\mathbb{R} \times \dots \mathbb{R}}_{q \text{ times}} \times \mathbb{N}~.$
The admissible transitions are of the form 
$v = (-e_j; 0)$ or $v =  (C e_j; -1)$, where $C \sim \mathrm{Unif}$, or $v = (\mathbf{0}; +1)$
These transitions are only admissible as long as $\rho_{i+1} \in \mathcal{S}$, which is considered fixed. 
Figure \ref{fig:setS} illustrates the setup with the constrain set $\mathcal{S}$. 
Since now the environment is stochastic, we can't define the same distance to a target goal in order to build the RL curriculum. Consequently, the first stage of the curriculum are goals with high probability of being at just one transition away from the start, and next stages are defined by the euclidean distance in the attribute space between the start and the end of the task. 
Similarly as with other enviroments, at test-time, the goals are to reach a certain point in attribute space. Since being close to the boundary reduces the probability of executing a transition, planning in this environment essentially consists in traversing the attribute map trying to stay away from the boundary of $\mathcal{S}$.
Table \ref{mazebaseresults} shows that our structured attribute model is able to leverage the geometric structure of the environment to significantly outperform both RL and unstructured attribute baselines. 




\section{Discussion}

This work is a first step towards mitigating the 
scalability issues of the unstructured attribute model from \cite{zhang2018composable}. However, there are still important limitations that we are planning to address in current/future work. First, we are currently extending the model to operate on larger, multiagent environments such as StarCraft, as well as 3d environments.
Next, currently each environment is given the correct arithmetic/geometric blocks, which is an unrealistic assumption in many real-life scenarios. The objective is to provide the agent with a large `dictionary' of such blocks (e.g. numerals, modulars, discrete symmetry groups, etc.), and learn how to select the appropriate structure for each attribute.  

\bibliography{references}
\bibliographystyle{plain} 

\appendix

\section{Further Experimental Setup}

\begin{figure*}
    \centering
    \includegraphics[width=0.3\textwidth,height=3cm]{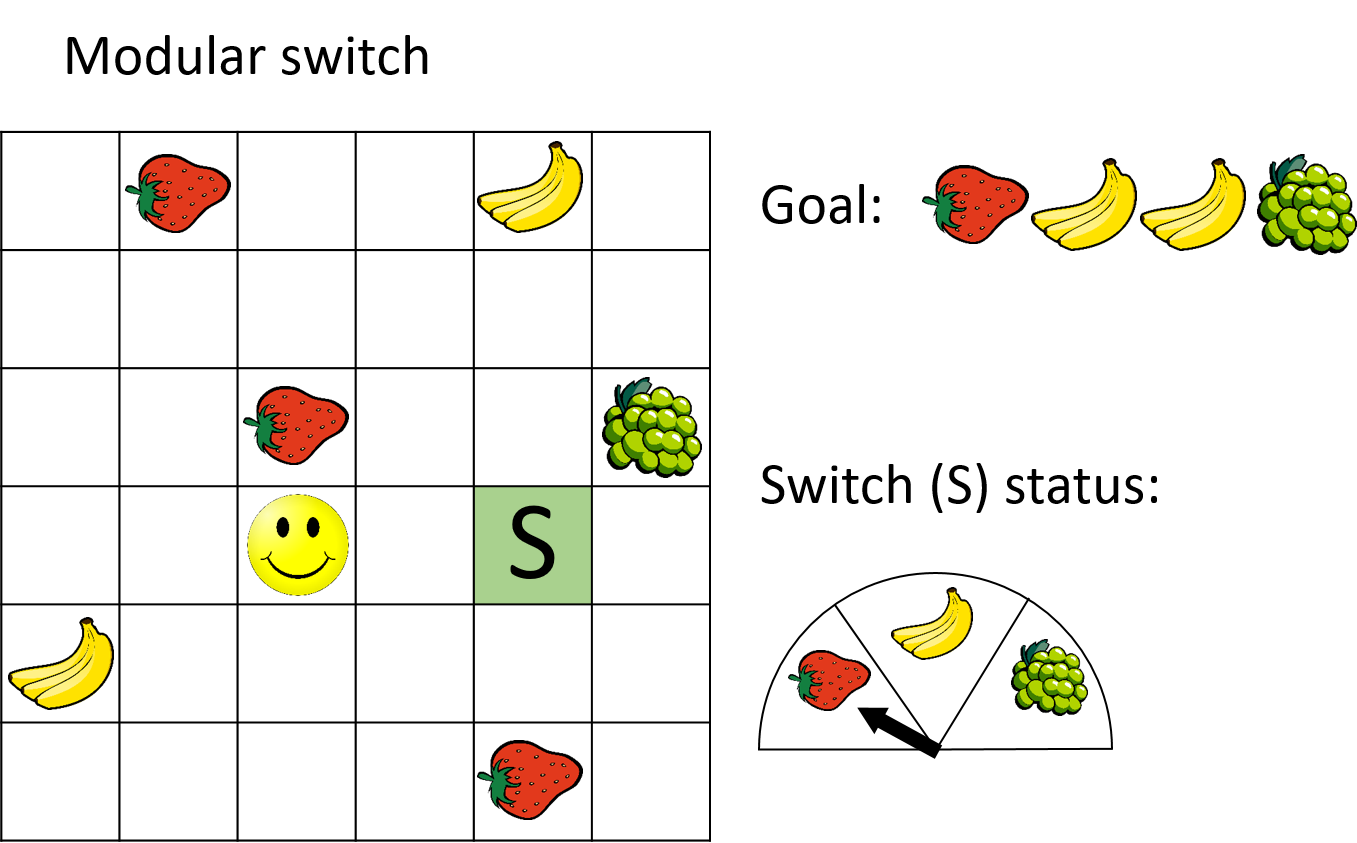}
    \includegraphics[width=0.3\textwidth,height=3cm]{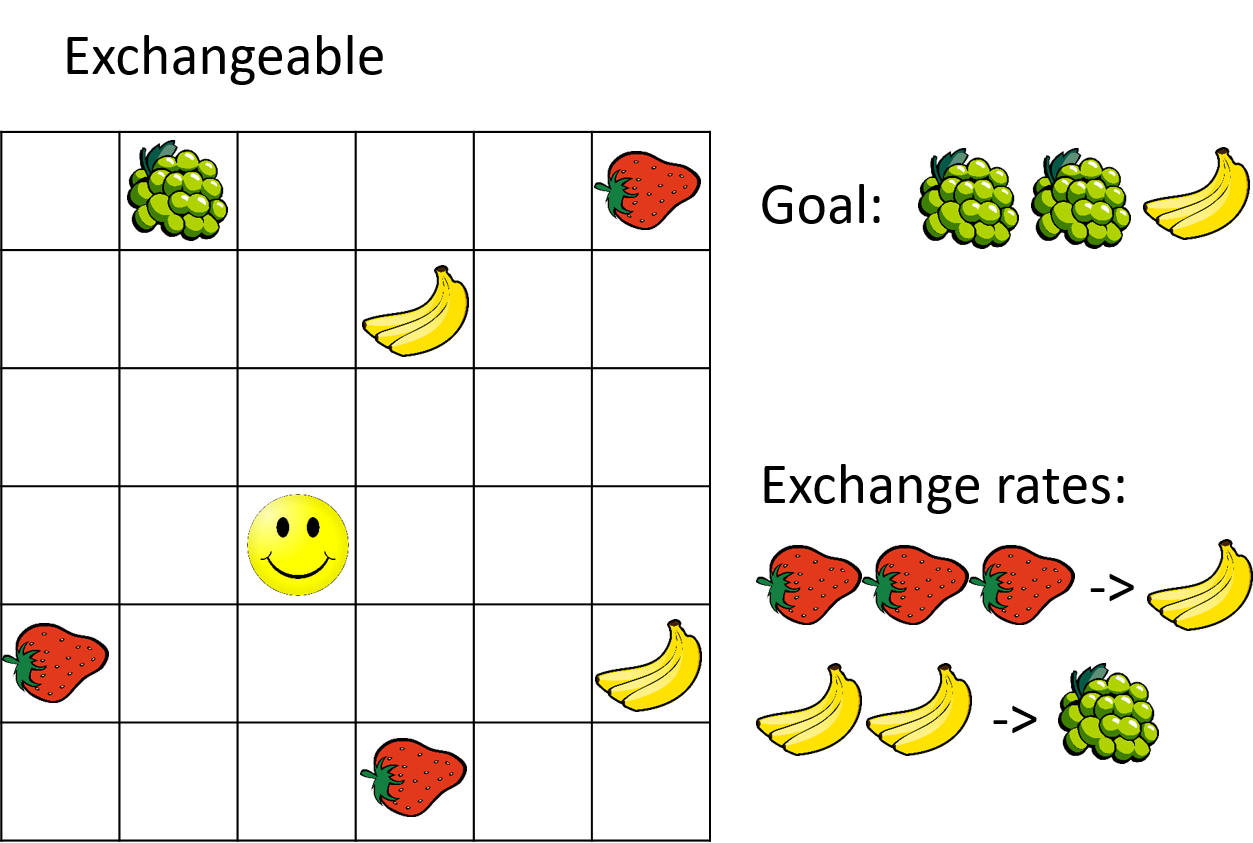}
    \includegraphics[width=0.3\textwidth,height=3cm]{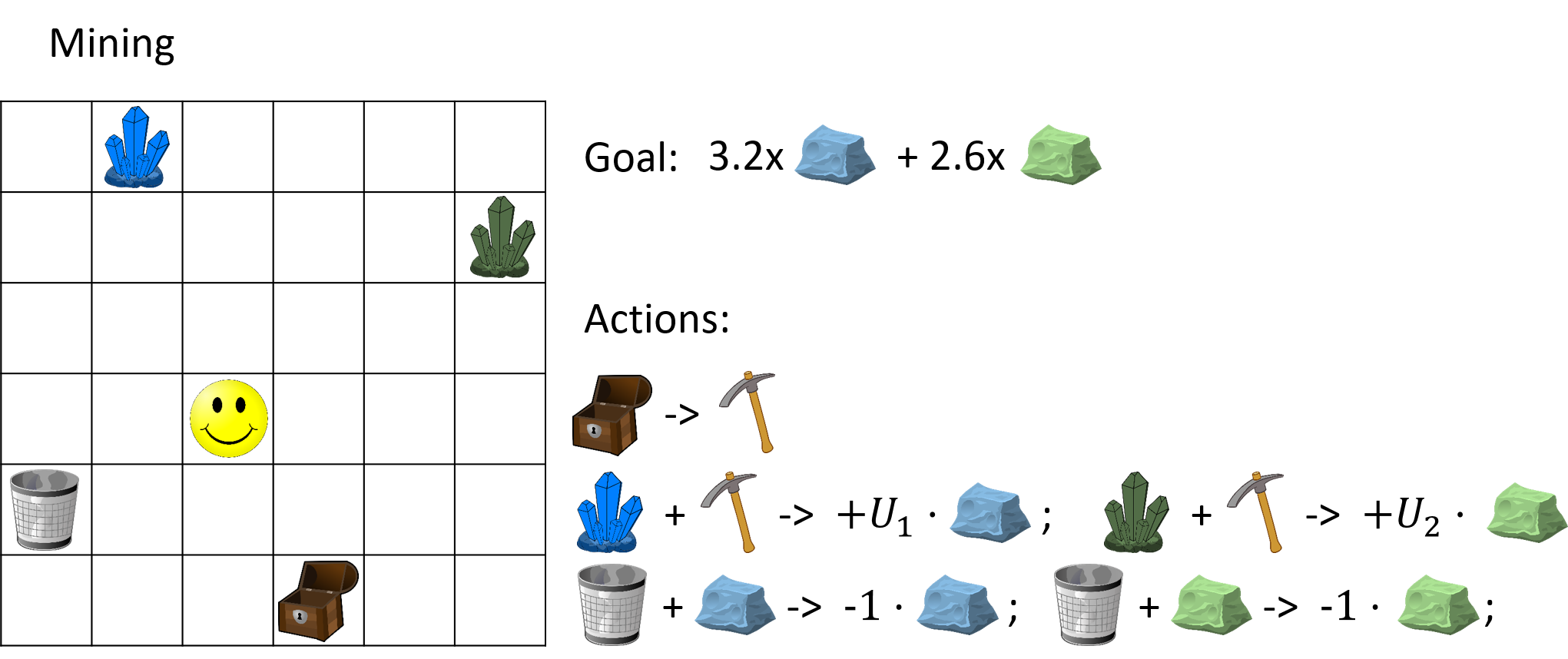}
    \caption{Schema of the grid environments. \textit{left:} Modular, \textit{center:} Exchangeable, \textit{right:} Mining}
    \label{fig:games}
\end{figure*}

\begin{figure*}
    \centering
    \includegraphics[width=0.3\textwidth]{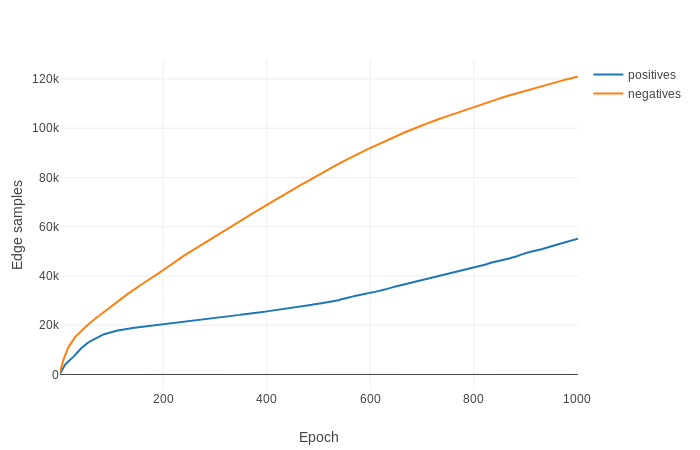}
    \includegraphics[width=0.3\textwidth]{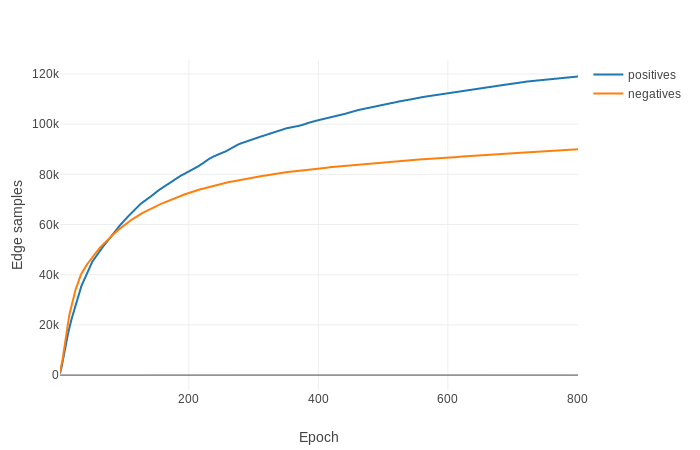}
    \includegraphics[width=0.3\textwidth]{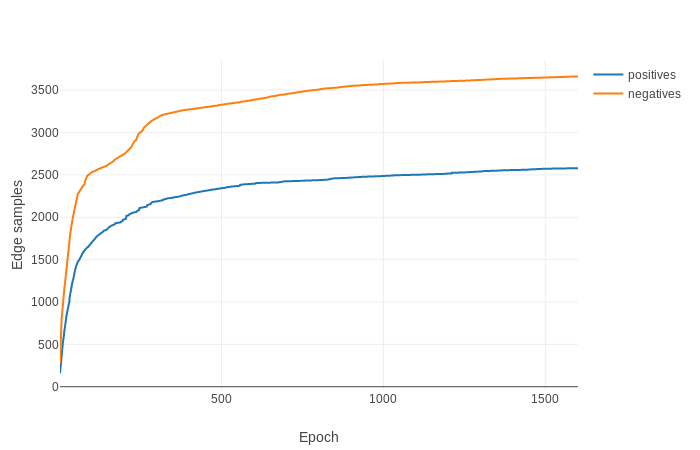}
    \caption{Rate at which positive and negative examples are being generated by $\piexp$ and $\piexe$ on the three environments (\textit{left:} Modular, \textit{center:} Exchangeable, \textit{right:} Mining)}
    \label{fig:labelcounts}
\end{figure*}

\begin{figure*}
    \centering
    \includegraphics[width=0.3\textwidth]{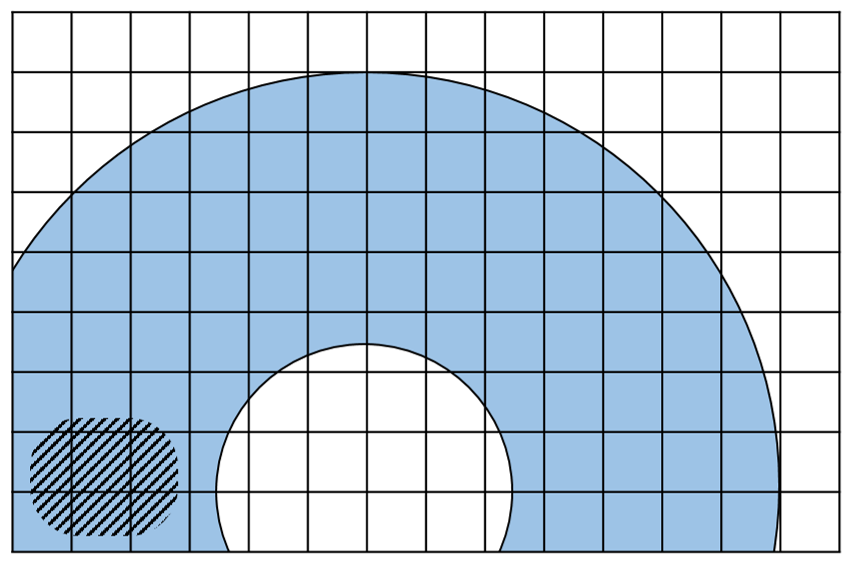}
    \caption{Support $\mathcal{S}$ of the attributes in the constrained attributes game. The black strips show the area where the game starts}
    \label{fig:setS}
\end{figure*}

\subsection{State-Attribute Regressor and Parametrization}

In the case where $f$ is not given by the user, on can train an estimator 
$\hat{f}$ from labeled pairs $\{ s_i, \rho_i\}_{i \leq I} \in (\mathcal{S} \times \RR)^I$, with a neural network trained with a mean-squared loss that reflects the geometry of each target 
attribute coordinate. If $\RR = G_1 \times \dots G_K$, $\rho=(\rho[1], \dots, \rho[K])$, 
and $y=(y[1], \dots, y[K])$ is the output of the neural net regressor, we consider the following metric on each $G_k$:
\begin{itemize}
    \item If $G_k = \N$, then $y[k] \in \R$, $\ell_k(y[k], \rho[k]) = | y[k] - \rho[k] |^2$ and $\hat{\rho[k]}=\lceil y[k] \rceil$.
    \item If $G_k = \R$, then $y[k] \in \R$, $\ell_k(y[k], \rho[k]) = | y[k] - \rho[k] |^2$ and $\hat{\rho[k]}= y[k] $.
    \item If $G_k = \Z/(q \Z)$, then $y[k] \in S^1$, $y[k] = \frac{u}{\|u\|}=(\sin \theta, \cos \theta)$, $\ell_k(y[k], \rho[k]) = 1-\langle y[k], e^{2\pi i \rho[k]/q}\rangle $ and $\hat{\rho[k]}= \lceil \theta q / 2\pi \rceil$.
\end{itemize}
The loss aggregated through all attribute coordinates becomes $\ell(y, \rho) = \sum_{k \leq K} \ell_k(y[k],\rho[k])$.

\subsection{Modular Switches}
For our model and all baselines we have used a two fully-connected layers net with 128 hidden units per layer. The batch size was of 5000 steps for the policies. We have trained the exploration policy with $\alpha_1=1$ and $\alpha_2=0$.

\subsection{Exchangeable Attributes}
We have used a two fully-connected layers net with 128 hidden units per layer. The batch size was of 5000 steps for the policies. We have trained the exploration policy with $\alpha_1=0$ and $\alpha_2=1$.
In this experiment we have made the rewards of the exploration continuous in time, in the sense that on each step we give reward not just for the transition that finishes the episode, but also for all the transitions that come after that episode until the end of the game. This way we are encouraging the explorer to look for trajectories that lead to late unseen attributes. This didn't work on the other experiments, because it stimulates the policy to do as much transitions as possible, and it got stuck in places where one could execute a number of transitions in a row (mostly the switch, but also the hammer store, the dump, etc).

\subsection{Constrained Attributes}
We have used a two fully-connected layers net with 128 hidden units per layer. The batch size was of 5000 steps for the policies. We have trained the exploration policy with $\alpha_1=0$ and $\alpha_2=1$.

In this game the transitions are only admissible as long as $\rho_{i+1}\in \mathcal{S}$. In our experiments we have defined this support as $\mathcal{S}=\{ v\in \mathbb{R}^+\times\mathbb{R}^+ \mid 2.5\le d(v,(6,1))\le 7\}$. This set is illustrated in Figure \ref{fig:setS} as the blue zone. In the figure, the black strips show the area where the agent starts the game.

\end{document}